\documentclass[a4paper]{article}

\usepackage{INTERSPEECH2022}
\usepackage{xcolor}
\usepackage{soul}
\usepackage{microtype}

\usepackage[noadjust]{cite}

\title{Analysis of Self-Supervised Learning and Dimensionality Reduction Methods in Clustering-Based Active Learning for Speech Emotion Recognition}
\name{Einari Vaaras$^1$, Manu Airaksinen$^2$, Okko Räsänen$^1$}
\address{
  $^1$Unit of Computing Sciences, Tampere University, Finland\\
  $^2$Helsinki University Hospital, Helsinki, Finland}
\email{einari.vaaras@tuni.fi, manu.airaksinen@hus.fi, okko.rasanen@tuni.fi}

\begin{document}

\maketitle
\begin{abstract}
When domain experts are needed to perform data annotation for complex machine-learning tasks, reducing annotation effort is crucial in order to cut down time and expenses. For cases when there are no annotations available, one approach is to utilize the structure of the feature space for clustering-based active learning (AL) methods. However, these methods are heavily dependent on how the samples are organized in the feature space and what distance metric is used. Unsupervised methods such as contrastive predictive coding (CPC) can potentially be used to learn organized feature spaces, but these methods typically create high-dimensional features which might be challenging for estimating data density. In this paper, we combine CPC and multiple dimensionality reduction methods in search of functioning practices for clustering-based AL. Our experiments for simulating speech emotion recognition system deployment show that both the local and global topology of the feature space can be successfully used for AL, and that CPC can be used to improve clustering-based AL performance over traditional signal features. Additionally, we observe that compressing data dimensionality does not harm AL performance substantially, and that 2-D feature representations achieved similar AL performance as higher-dimensional representations when the number of annotations is not very low.
\end{abstract}

\section{Introduction}

In many complex real-world machine-learning applications, annotating data can be expensive and time-consuming. This is often the case particularly in situations where domain experts are needed to carry out the annotation process \cite{al_cheap_and_fast, human_in_the_loop_for_ml}. For such cases, active learning (AL) algorithms can be used to reduce human annotation effort and to produce machine-learning models that perform well with limited labeled data \cite{al_in_practice_paper}. For example, data scarcity is an ever-present problem in deploying speech emotion recognition (SER) systems to new domains \cite{ser_al_sparse}. %\okk{\st{when developing new SER-based machine-learning systems \cite{ser_al_sparse}.}}

There are a number of different approaches for AL, of which by far the most popular methods are based on uncertainty or confidence scores of a classifier trained on already-labeled data (e.g. \cite{ser_al_sparse, al_for_dimensional_ser, al_asr, al_for_sound_classification, al_adversarial_sampling}). However, when the maximum number of labels that can be manually assigned, also known as the \textit{labeling budget}, adds up to only a small subset of the data, the aforementioned AL approaches cannot be applied. This is because these methods often require a rather large number of annotated samples before they can outperform random sampling \cite{al_difficulties}.

In the absence of any existing annotated data, a potential approach to AL is to utilize the distributional properties of the dataset with clustering-based AL methods (e.g. \cite{al_pre_clustering, al_with_clustering, al_low_resource_sr, al_subspace_clustering}). These methods rely heavily on how the samples are organized in the feature space (i.e. the choice of features) and what distance metric is used, as the methods need to use these two to cluster the data points and to prioritize the order in which cluster samples are provided for human annotators. This puts particular emphasis on how the features behave in the given metric space with respect to the analysis task at hand. In addition, the dimensionality of the used features poses a potential challenge for the AL algorithms, as accurate data density estimation from finite data becomes less accurate in higher-dimensional spaces \cite{high_dim_density_estimation}. Hence, it could be beneficial for AL if there was a systematic way of representing data samples in a low-dimensional feature space. Moreover, a lower dimensionality would heavily reduce the computational complexity of AL algorithms. Taken to the extreme, a 2-D feature space could be used to combine AL with data visualization and efficient human-based data exploration and reorganization, such as in the annotation platform described in \cite{kth_paper}.

\begin{figure}[b]
    \centering
    \vspace{-6 pt}
    \includegraphics[width=0.3\textwidth]{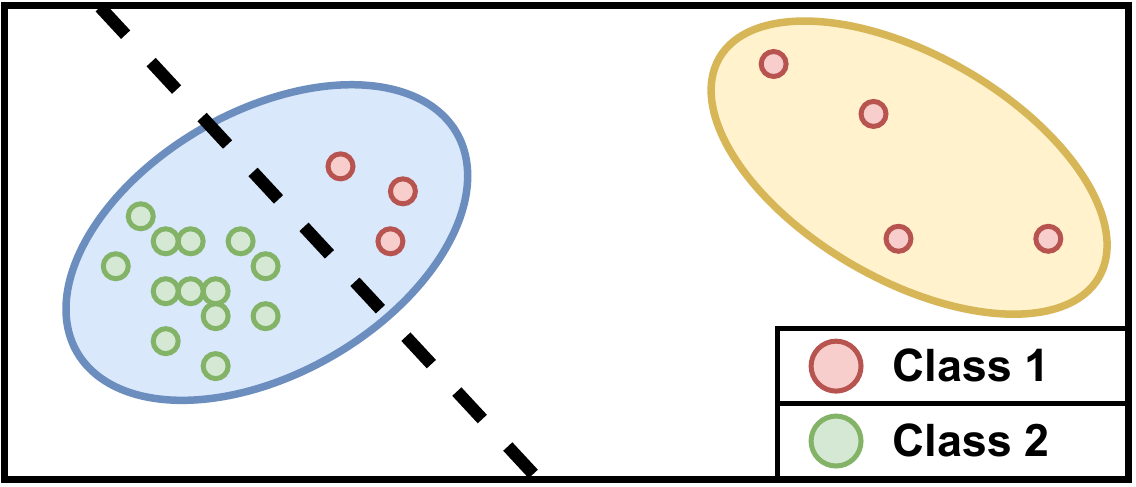}
    \vspace{-6 pt}
    \caption{A toy example for 2-D data, clustered into two clusters (blue and yellow) with $k$-means clustering using Euclidean distance. Although the classes are linearly separable (dashed line), the data points are not organized into class-specific clusters.}
    \label{fig_toy_example}
\end{figure}

While the use of standard acoustic features such as i-vectors \cite{al_low_resource_sr} or MFCCs \cite{tuni_mal} has been applied to previous AL clustering approaches, there are nowadays a number of highly promising unsupervised methods for learning feature spaces (e.g. \cite{cpc, trill_paper, data2vec_paper}). These methods can learn linearly separable representations for many speech phenomena of interest (e.g. \cite{cpc}), and could potentially be used to learn organized feature spaces for clustering-based AL algorithms. However, the aforementioned methods typically result in high-dimensional feature spaces which can impose challenges for AL algorithms, as discussed above. Also, these methods have been used primarily for learning features that are relevant for supervised downstream tasks, but not for intermediate tasks such as clustering, where data grouping is equally important to data separability (see Fig. \ref{fig_toy_example} for an example). %As an example, contrastive predictive coding (CPC) \cite{cpc} aims to produce linearly separable features but, as shown in Fig. \ref{fig_toy_example}, linear separability does not necessarily mean good clusterability.

In the present study, we combine AL with unsupervised learning and dimensionality reduction methods in order to seek answers to the questions of whether self-supervised representation learning (SSRL) can improve AL performance over classical signal features, and how AL performance depends on input feature dimensionality. While doing so, the work also sheds light on the question of how the local and global topology of a feature space affects clustering-based AL systems.
%Is the local structure of high-dimensional features enough for clustering-based AL approaches, or is the global structure needed for these approaches to work? How much can we compress the feature dimensionality without losing AL performance? Can CPC be utilized to systematically create a clusterable feature space? 
In order to answer these questions, we conduct a set of experiments involving a simulated annotation procedure and a clustering-based AL algorithm from \cite{tuni_mal}. We use SER as our test case using four different SER corpora and with two distinct classification tasks. Our ultimate goal is to find functioning practices for developing machine-learning algorithms for applications where access to human expert labels is expensive. Hence, our primary focus is in cases with a low labeling budget, although we also include higher labeling budgets in our present experiments.

\section{Methods}

\begin{figure}[h]
    \centering
    \includegraphics[width=0.47\textwidth]{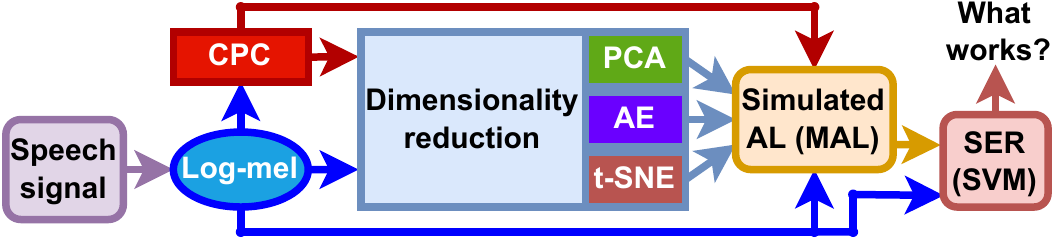}
    \vspace{-6 pt}
    \caption{A block diagram of the present experimental setup.}
    \vspace{-6 pt}
    \label{fig_block_diagram}
\end{figure}

Fig. \ref{fig_block_diagram} depicts a block diagram of the present experiments. First, a standard acoustic feature (log-mel) representation is obtained from an input signal. Then, an SSRL algorithm called contrastive predictive coding (CPC) \cite{cpc} is applied to the data. CPC aims to produce linearly separable features that can be used to predict signal evolution over time \cite{cpc_image_recognition, cpc_linear_separability}. CPC has already been successfully used with clustering-based approaches (e.g. \cite{cpc_linear_separability, takahashi21_interspeech, maekaku21_interspeech}) and also produces features that separate suprasegmental properties such as speaker identities \cite{cpc}. However, to the best of our knowledge, \cite{active_contrastive_learning} is the only study so far using CPC for AL. 

Next, an array of alternative dimensionality reduction methods is optionally applied to the log-mel and CPC features. %If the features have low intrinsic dimensionality, then \textit{manifold learning} methods such as the t-distributed stochastic neighbor embedding (t-SNE) \cite{tsne} can be used for reducing feature dimensionality. t-SNE preserves the local structure of the high-dimensional data while also revealing some global aspects, such as clusters at multiple scales \cite{tsne}. 
In the present study, we explore the use of t-distributed stochastic neighbor embedding (t-SNE) \cite{tsne}, nonlinear bottleneck autoencoders (AEs), and principal component analysis (PCA) for dimensionality reduction. Among these, t-SNE preserves the local structure of the high-dimensional data while also revealing some global aspects, such as clusters at multiple scales \cite{tsne}. In contrast, PCA simply maps the data into principal axes of variation without differentially altering the local and global metric structure of the feature space. Bottleneck AEs, on the other hand, attempt to learn a low-dimensional feature embedding from which the original input features can be reconstructed.

Finally, we run a simulated AL-based speech emotion annotation and a SER classifier deployment procedure using a support vector machine (SVM) classifier, similar to \cite{einari_dippa}. For this, we use a %with a 
clustering-based AL algorithm %using 
together with each possible combination of log-mel or CPC features and the aforementioned dimensionality reduction methods, including not using dimensionality reduction at all, and compare the resulting SER performance among the alternative strategies. %We use medoid-based active learning (MAL) \cite{tuni_mal} as our AL algorithm, which is an AL method developed for scarce labeling budgets.

\subsection{Contrastive predictive coding}

CPC \cite{cpc} is an unsupervised method for extracting representations from data that encode underlying shared information between different parts of the input signal. This is achieved by predicting $k \in \{1, ..., K\}$ future latent representations of the input signal where typically $K > 1$ (e.g. $K = 12$ for speech in \cite{cpc}). A CPC model consists of two separate models, a nonlinear encoder, $g_{enc}$, and an autoregressive model, $g_{ar}$. First, $g_{enc}$ maps the input observations, $\bm{x}_t$, into latent representations $\bm{z}_t = g_{enc}(\bm{x}_t)$. Then, $g_{ar}$ maps $\bm{z}_{\le t}$ into a context latent representation $\bm{c}_t = g_{ar}(\bm{z}_{\le t})$. Instead of directly modeling $\bm{x}_{t+k}$, CPC models a density ratio that aims to preserve the mutual information between $\bm{x}_{t+k}$ and $\bm{c}_t$. For this, a log-bilinear model $f_k(\bm{x}_{t+k}, \bm{c}_t) = exp\left(\bm{z}_{t+k}^T W_k \bm{c}_t\right)$ is used,
%
%\begin{equation} \label{log_bilinear_model}
%    f_k(\bm{x}_{t+k}, \bm{c}_t) = exp\left(\bm{z}_{t+k}^T W_k \bm{c}_t\right) \ ,
%\end{equation}
%
where $W_k$ are linear transformations. Both $g_{enc}$ and $g_{ar}$ are trained using the loss

\begin{equation} \label{info_nce}
    L_{\text{CPC}} = -\frac{1}{K} \sum\limits_{k=1}^{K} log \left[\frac{f_k(\bm{x}_{t+k}, \bm{c}_t)}{\sum\limits_{\bm{x}_j\in X}{f_k(\bm{x}_j, \bm{c}_t)}}\right] \ ,
\end{equation}
where $X = \{\bm{x}_1, ..., \bm{x}_N\}$ is a set of $N$ random samples containing one positive sample and $N-1$ negative samples. As shown in \cite{cpc}, minimizing the loss in Eq. \ref{info_nce} maximizes the mutual information between $\bm{x}_{t+k}$ and $\bm{c}_t$.

\subsection{Medoid-based active learning}

For AL, we use medoid-based active learning (MAL) \cite{tuni_mal}, which is an AL method developed for scarce labeling budgets.
%\okk{\st{The MAL \cite{tuni_mal} 
The algorithm consists of three subsequent stages: 1) compute an affinity matrix containing the pairwise distances between each sample in a dataset, 2) perform $k$-medoids clustering using this affinity matrix, and 3) in a descending cluster size order, query for human annotations for the medoids.

In the first stage, the pairwise distances between each sample in a dataset are computed using a distance metric, $d$, and are stored to an affinity matrix, $A$. In the second stage, $k$-medoids clustering (see e.g. \cite{k_medoids_algorithm}) is performed for the data using $A$. First, one sample is randomly selected as a member of a set, $S$. Then, $k-1$ additional samples are added to $S$ one at a time using the \textit{farthest-first traversal} algorithm \cite{farthest_first_traversal}, each added sample being farthest from the current set $S$. Here, the distance from a sample, $\bm{a}$, to the set $S$ is defined as $d(\bm{a},S) = \underset{\bm{b}\in S}{\textrm{min}} \ d(\bm{a},\bm{b})$.
%
%\begin{equation} \label{eq_farthest_first}
%    d(\bm{a},S) = \underset{\bm{b}\in S}{\textrm{min}} \ d(\bm{a},\bm{b}) \ .
%\end{equation}
%
Next, the samples in $S$ are used as initial medoids for a $k$-medoids clustering algorithm in order to assign each sample into one of the $k$ clusters. Based on a previous study \cite{vaaras21_interspeech}, $k$ was set to be $\frac{N}{3}$, where $N$ is the number of samples in a corpus.% This can also be interpreted as meaning that the average cluster size is three.

In the third stage, the clusters are first sorted in a descending-size order, after which the cluster medoids are presented to human annotators for labeling. In the present experiments, the obtained labels were used in two different ways: i) using only the medoid labels as labeled data (referred to as ``medoid labels"), or ii) propagating the medoid label to all elements in the cluster (referred to as ``cluster labels").

\section{Experiments}

\subsection{Emotional Speech Corpora} \label{sec_ser_corpora}

We conducted our experiments using four SER corpora:

1) \textit{The Berlin Emotional Speech Database} (EMO-DB) \cite{emodb} is perhaps the most widely used SER corpus. It contains 535 spoken utterances in German from 10 actors in seven emotions: anger, boredom, disgust, fear, joy, neutral, and sadness.
    
2) \textit{eNTERFACE} \cite{enterface} contains 1,287 videos in English (42 test subjects, 14 nationalities), of which only the audio tracks were used in the present study. The corpus contains emotions in six categories: anger, disgust, fear, joy, sadness, and surprise.
    
3) \textit{The Finnish Emotional Speech Corpus} (FESC) \cite{finnish_corpus} consists of 450 spoken passages from nine Finnish professional actors portraying five emotions: neutral, sadness, joy, anger, and tenderness. Based on long silences defined by an energy threshold \cite{einari_dippa}, the passages were further split into 4,254 utterances.
    
4) \textit{The Ryerson Audio-Visual Database of Emotional Speech and Song} (RAVDESS) \cite{ravdess} is a multimodal database containing 7,356 recordings in English from 24 professional actors. Only the recordings including speech (1,440 utterances) were used in the present experiments, including eight emotions: neutral, calm, happy, sad, angry, fearful, surprise, and disgust.

In order to harmonize the emotional labels of each SER corpus, the labels were mapped into the quarters of the valence-arousal plane following the mapping of \cite{schuller_cross_corpus_ser}, which has been popularly used in SER studies (e.g. \cite{ser_unite_or_vote, ser_cross_lingual, ser_emotion_discriminative_features}). The mapping also simplifies the SER classification task into two binary classification tasks: valence (positive/negative) and arousal (high/low).

\begin{figure}[t]
    \centering
    \vspace{-12 pt}
    \includegraphics[width=0.33\textwidth]{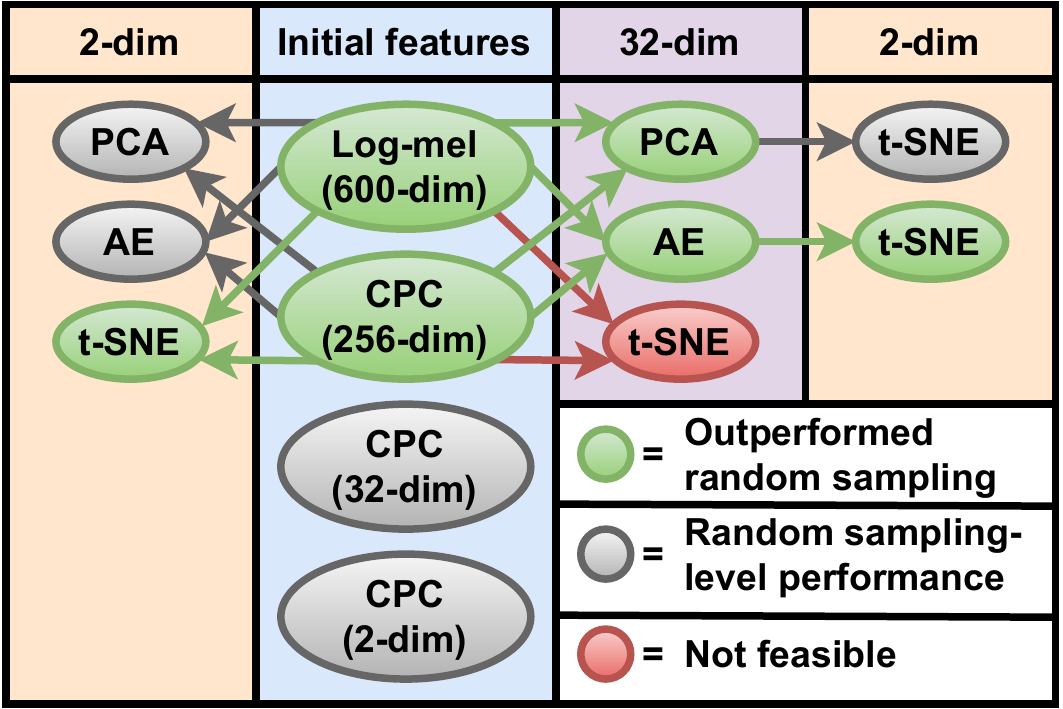}
    \vspace{-6 pt}
    \caption{A visualization of the feature, dimensionality reduction method, and data dimensionality combinations that were experimented with in the present study.}
    \label{fig_tested_combinations}
    \vspace{-12 pt}
\end{figure}

\subsection{CPC and AE model training} \label{cpc_ae_model_training}

A separate CPC model was trained for each SER corpus as a data-driven feature extractor. Log-mel frames (40 mel filters, 30-ms Hann window, 10-ms shifts) were used as input features for the encoder $g_{enc}$. To get constant-length inputs, 5-second segments were extracted from the acoustic signal by zero-padding utterances shorter than 5 s and randomly selecting 5-second segments of utterances longer than 5 s. The encoder $g_{enc}$ consisted of three fully-connected ELU \cite{elu_original} layers of 256 units, each followed by a dropout of 20\%. The autoregressive model $g_{ar}$ was a one-layer GRU \cite{gru_original} with a 256-dimensional hidden state. %As in \cite{cpc}, a linear transformation was used for $W_k$.
Prediction was carried up to 12 steps (120 ms) ahead.

Each corpus was randomly split into training and validation sets in a ratio of 80:20 utterances. The models were trained using the loss $L_{\text{CPC}}$ in Eq. \ref{info_nce}, a batch size of 8 utterances, and Adam \cite{adam_original} optimizer. An initial learning rate of $10^{-4}$ was used with a reduction factor of 0.7 based on the validation loss with a patience of 20 epochs. Early stopping with a patience of 100 based on validation loss was used to select the model with the lowest validation loss. For each sample in a minibatch, the rest of the minibatch samples acted as the negative samples.

Using a similar 80:20 split as above, an AE model was trained for dimensionality reduction individually for each corpus and for two input features (utterance-level log-mel and CPC features, see Sec. \ref{sec_features}). The network consisted of six fully-connected ELU layers of 512 units each, except for the third layer (32 units) and the last layer (600 and 256 units for log-mel and CPC features, respectively). The 512-unit layers had a dropout of 10\%. The AE models were trained using MSE loss, Adam optimizer, batch size of 1024, and a learning rate of $10^{-4}$. Early stopping with a patience of 300 based on validation loss was used, and the best encoder (the first three network layers) was selected from the AE model with the lowest validation loss.

\begin{figure}[t]
    \centering
    \vspace{-12 pt}
    \includegraphics[width=0.43\textwidth]{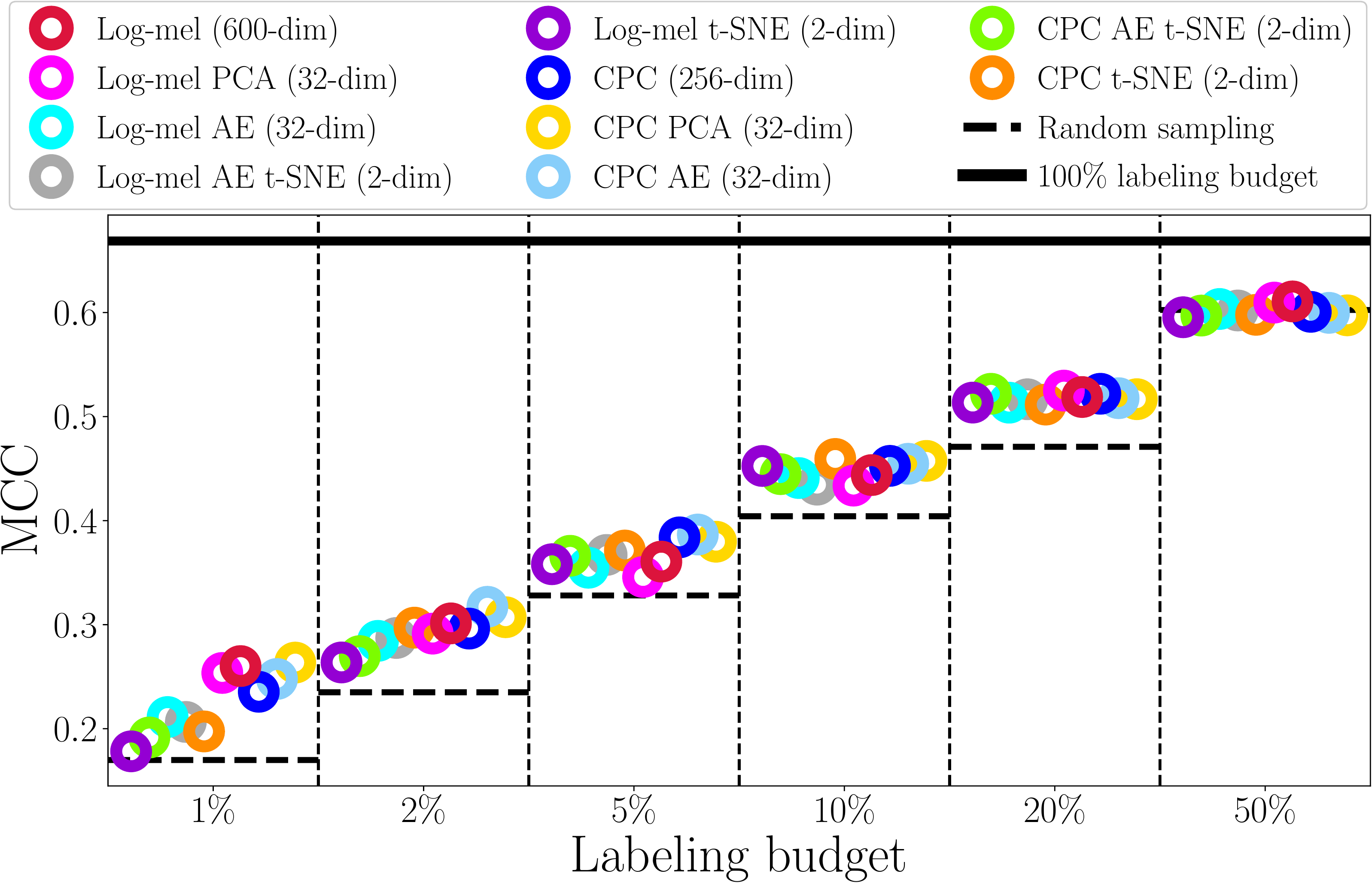}
    \vspace{-6 pt}
    \caption{The MCC performance scores for different experimented features using the MAL algorithm and varying labeling budget, and with random sampling as a baseline reference. Each reported number is the mean of 200 feature-specific experiments (valence and arousal classification tasks, four SER corpora, 5-fold CV, five different random initializations for each data split). The features are ordered based on mean performance, with the best-performing feature being the rightmost.}
    \vspace{-6 pt}
    \label{fig_test_bench_results}
\end{figure}

\begin{figure*}[t]
    \centering
    \vspace{-12 pt}
    \includegraphics[width=0.43\textwidth]{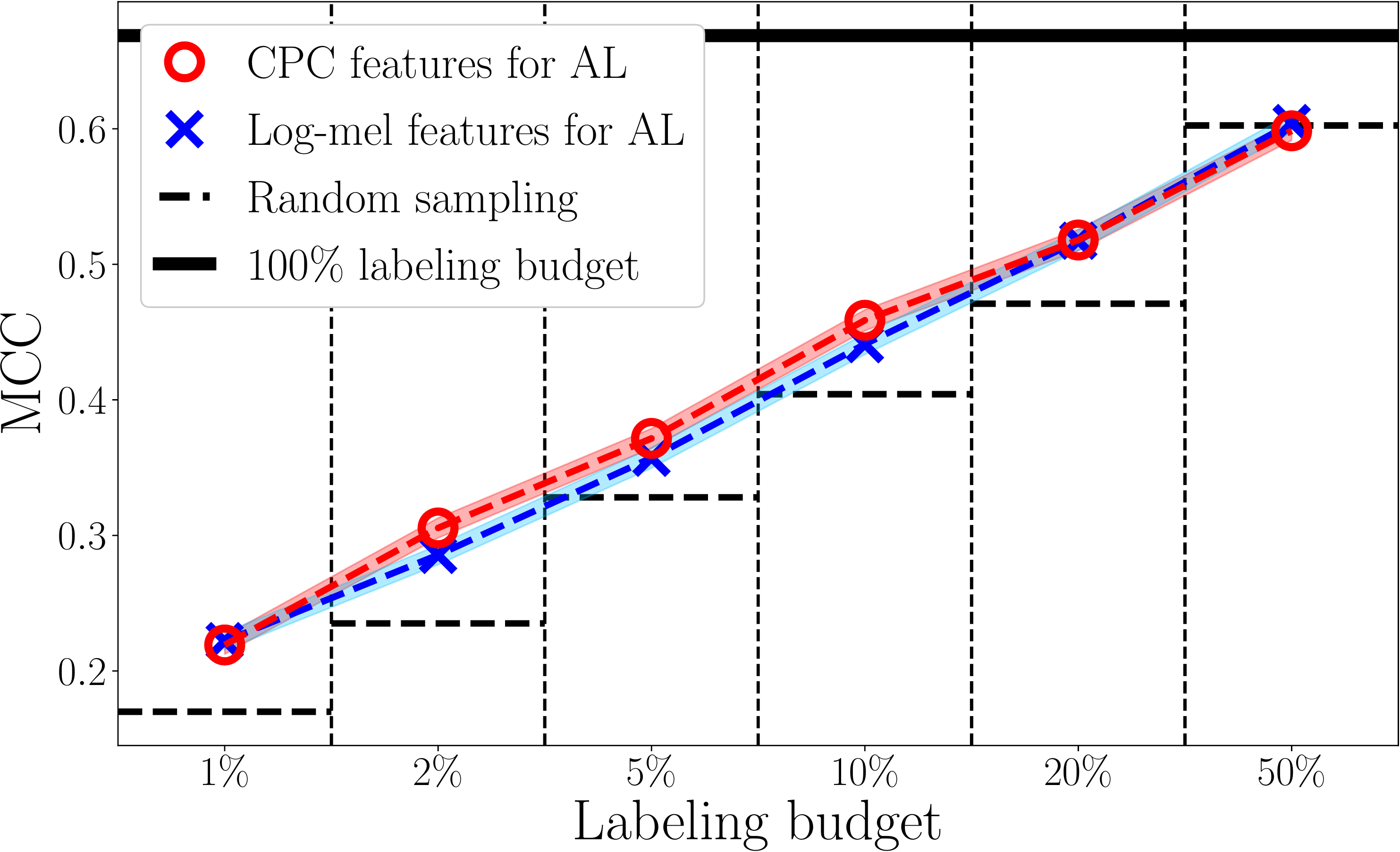}
    \hspace{16pt}
    \includegraphics[width=0.43\textwidth]{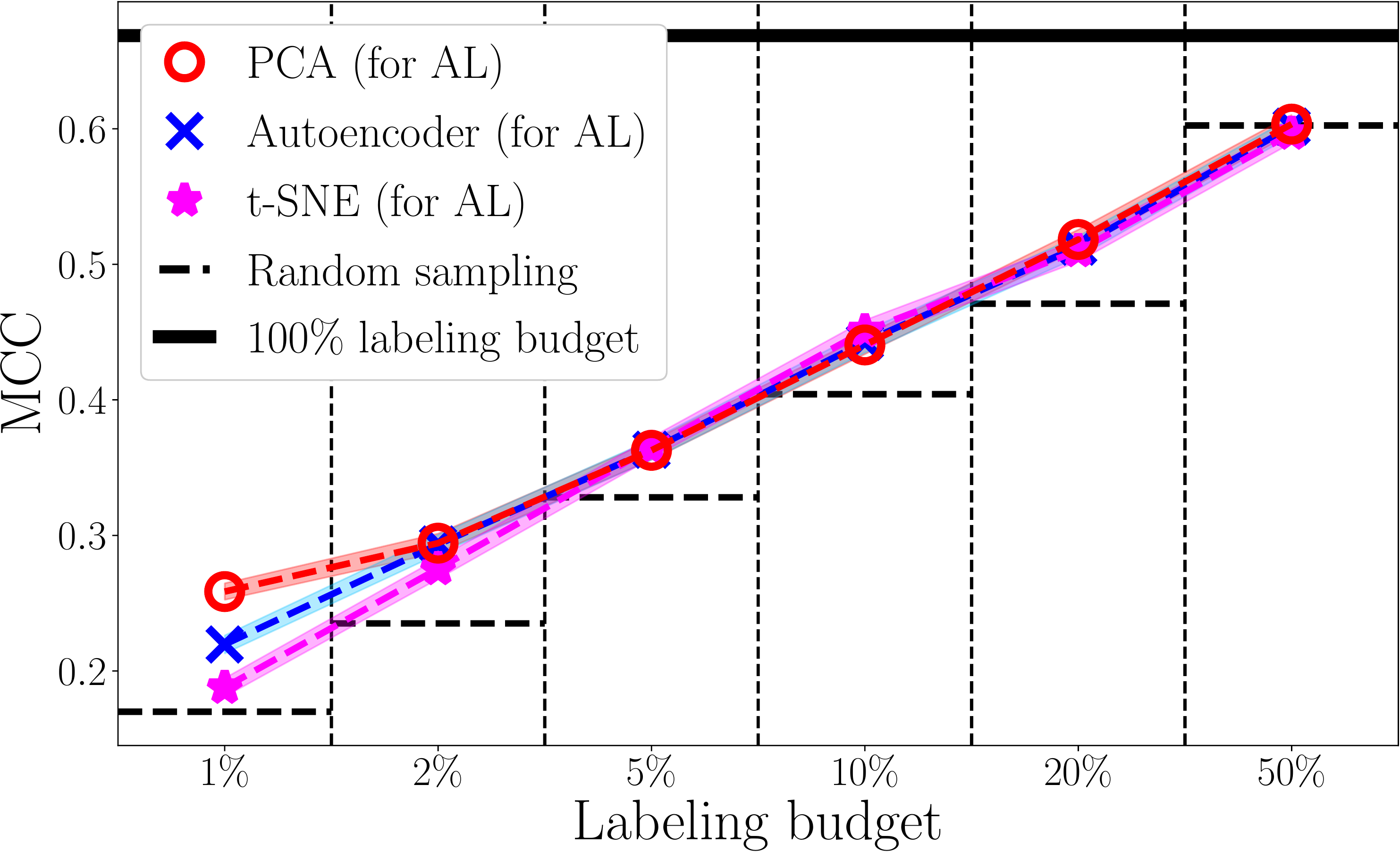}
    \caption{Comparison of CPC and log-mel features (left) and PCA, AE, and t-SNE dimensionality reduction methods (right) in MAL-based SER simulations, averaged across all experimental conditions (standard error as shaded area). For the latter, the cases where only one dimensionality reduction method is being used are considered. Random sampling is shown as a baseline reference. Note that a SER classifier is always trained using 600-dim utterance-level log-mel features independently of the features and dimensionality used for AL.}
    \vspace{-4 pt}
    \label{fig_comparisons}
\end{figure*}

\subsection{Features} \label{sec_features}

For log-mel features of each utterance, seven functionals (the first four moments, min, max, and range) were applied to the time dimension of log-mel frames (Sec. \ref{cpc_ae_model_training}), and the first four moments were applied to the first and second order delta features to get a 600-dimensional utterance-level feature vector. For CPC features, the mean over the time dimension of the encoder outputs $\bm{z}_t$ was computed to get 256-dimensional utterance features (ignoring potential zero padding). The context latent representations $\bm{c}_t$ were also tested but resulted in slightly worse results than using $\bm{z}_t$, and hence are not separately reported. In addition to the mean, we also tested computing other functionals from the CPC features. However, since the mean is the only moment that preserves the metric properties of the original CPC feature space, the inclusion of these additional functionals turned out to systematically worsen the SER performance of the CPC features, and hence are also not reported.

Fig. \ref{fig_tested_combinations} illustrates the tested combinations of features and dimensionality reduction methods that were included in the experiments. For both the 600-dimensional log-mel and 256-dimensional CPC features, 32- and 2-dimensional PCA and AE representations were computed. The AE representations were obtained using the encoder networks of Sec. \ref{cpc_ae_model_training}. Additionally, 2-D t-SNE features were computed from the high-dimensional log-mel and CPC features as well as their 32-dimensional representations. The Scikit-learn \cite{sklearn_reference} implementation of t-SNE was used with default values, with the exception of using PCA initialization for better overall stability. We also tested training CPC models directly into lower-dimensional feature spaces but with poor results. In order to save space, only the features that outperformed the random sampling baseline in AL-based classification are included in the present results. Computing 32-dimensional t-SNE features was not considered due to the poor scalability of the algorithm for higher than 2-dimensional projections.

\subsection{AL simulation setup for SER}

For each corpus and for each feature described in Sec. \ref{sec_features}, SER experiments were carried out with the MAL algorithm using a simulated annotation procedure. In the simulation, samples selected for annotation by MAL were replaced by the ground-truth labels of the corpus as available based on earlier human annotations, and were then used to train a classifier for the valence and arousal classification tasks (see also \cite{einari_dippa}). The data was randomly split into training and test sets using 5-fold cross-validation (CV), after which MAL was applied to the training set. Labeling budgets of 1\%, 2\%, 5\%, 10\%, 20\%, 50\%, and 100\% of the total training set samples were used in the experiments. Although MAL was originally intended for scarce labeling budgets, higher labeling budgets are also included in the analyses. Each cross-fold experiment was repeated five times to account for the minor variability in the results due to t-SNE and MAL random initialization. We use Matthew's correlation coefficient (MCC) \cite{mcc_reference} as our primary evaluation metric. As argued by e.g. \cite{mcc_argument}, MCC can be considered as one of the most informative quality measures for binary classifiers since it requires good results from all four confusion matrix categories to output a high score.

After obtaining a set of labeled data from MAL, an SVM with an RBF kernel was trained with the labeled training data and tested on the test data. This was applied to both classification tasks and for both medoid and cluster labels. Since the primary focus of the present experiments was to find features that enhance the performance of clustering-based AL algorithms, the training and testing features for the SVMs were standardized for each experiment. For this, we selected the 600-dimensional utterance-level log-mel feature statistics (z-score normalized at the corpus level), as they were found generally well-performing in valence and arousal classification tasks in \cite{einari_dippa}. SVM hyperparameters (box constraint and kernel scale parameter) were optimized separately for each corpus and classification task based on a grid search using 5-fold CV on all labeled data for the given corpus. As a baseline reference, random sampling results with corresponding labeling budgets are also reported, with experiments conducted in a similar manner as with MAL.

%\begin{figure}[t]
%    \centering
%    \includegraphics[width=0.43\textwidth]{cpc_vs_logmel.pdf}
%    \caption{Comparison of CPC and log-mel features in MAL-based SER simulations. Mean values (dashed lines) and standard error bars are shown across all experimental conditions. Random sampling is shown as a baseline reference.}
%    \label{fig_cpc_vs_logmel}
%\end{figure}
%

%As a baseline reference, random sampling \okk{results with the corresponding labeling budget is also reported}. %was used similar to the simulated annotation procedure with MAL. 
%The data was split into training and test sets using 5-fold CV, and $n \in [1, 2, 5, 10, 20, 50, 100]$\% of the labeled samples in the training set were randomly selected for training an SVM model, followed by testing the model on the test data. Each experiment was repeated five times with a different randomization for a total of 25 experiments altogether for each tested corpus/feature/labeling budget combination. 

For the sake of brevity, the results with MAL cluster labels are omitted since they provided an advantage over medoid labels only for 1\% labeling budget. For the same reason, only the results from the best-performing distance metrics for MAL are reported, corresponding to the Euclidean distance for 2-D features, and the cosine distance for higher-dimensional features.

%\begin{figure}[t]
%    \centering
%    \includegraphics[width=0.43\textwidth]{pca_vs_ae_vs_tsne.pdf}
%    \caption{Comparison of PCA, AE, and t-SNE dimensionality reduction methods in MAL-based SER simulations. Mean values (dashed lines) and standard error bars are shown across all experimental conditions where only one dimensionality reduction method is being used. Random sampling is shown as a baseline reference.}
%    \label{fig_pca_vs_ae_vs_tsne}
%\end{figure}
%

\section{Results}

%\begin{figure}[h]
%    \centering
%    \includegraphics[width=0.43\textwidth]{600dim_vs_32dim_vs_2dim.pdf}
%    \caption{Comparison of the original high-dimensional (600/256-dimensional), 32-dimensional, and 2-D feature spaces for the MAL algorithm. The reported results are the mean and SEM values of the dimension-related experiments pooled together, with random sampling as a baseline reference.}
%    \label{fig_600dim_vs_32dim_vs_2dim}
%\end{figure}
%

Fig. \ref{fig_test_bench_results} presents the results of the experimental setup, averaged over the corpora and both classification tasks (valence and arousal). As the labeling budget increases, the differences between the tested features become smaller. Also, with a 50\% labeling budget MAL does not provide a benefit over random sampling. For labeling budgets of 1\%--20\%, the 32-dimensional ``CPC PCA" features had the highest average performance (approx. 0.38 MCC). For the four SER corpora, a labeling budget of 1\% corresponds to approx. 15 samples on average.% Using 2-D t-SNE features computed from 256-dimensional CPC features provided only a minor drop in performance compared to the 256-dimensional features (approx. 0.37 MCC vs. 0.36 MCC mean performance with labeling budgets 1\%--20\%).

Fig. \ref{fig_comparisons} (left) shows the performance comparison between CPC and log-mel features. For labeling budgets of 2\%--10\%, CPC features provided statistically significant improvements over log-mel features (paired \textit{t}-test, $p < 0.05$, $df = 999$, $t = 3.06\textup{--}6.12$ across the tests), showcasing better overall clusterability and/or more efficient feature space exploration for the CPC features in the SER task. Fig. \ref{fig_comparisons} (right) shows the comparison between dimensionality reduction methods, demonstrating a similar performance for all three methods from a 5\% labeling budget onward. With a 2\% labeling budget, PCA and AE obtain a similar performance and outperform t-SNE, and with a 1\% labeling budget, PCA outperforms the AE method which, in turn, outperforms t-SNE (paired \textit{t}-test, $p < 0.05$, $df=399$, $t = 1.17\textup{--}3.92$ across the tests). Comparing different dimensionalities in Fig. \ref{fig_test_bench_results}, we can observe that heavily reducing the dimensionality of the 600/256-dimensional features does not provide a major drop in average performance, with ``CPC PCA" and ``CPC AE" even outperforming the 256-dimensional CPC features. The 600/256-dimensional and 32-dimensional features obtained similar performance on average, and the 2-D t-SNE features also obtained similar performance for labeling budgets of 5\% and larger. Overall, these results indicate that for clustering-based AL approaches, both the global (600/256/32-dimensional features) and local (t-SNE features) properties of the data can be used to provide an advantage over random sampling. However, it seems that t-SNE works best if the labeling budget is not below 5\%, possibly since t-SNE distorts the global distance metrics.

\section{Conclusions}

In this study, we combined CPC and various dimensionality reduction methods in search of functioning practices for clustering-based AL. Our experiments revealed that SSRL can be utilized to improve clustering-based AL performance over traditional signal features. In addition, we found that both the local and global topologies of feature spaces can be successfully used for AL. Furthermore, we observed that compressing the dimensionality of high-dimensional features does not provide a major drop in AL performance. We also found that 2-D t-SNE features achieved similar AL performance as higher-dimensional features when the labeling budget is not very low. This finding could be utilized to reduce the computational complexity of AL algorithms, and to combine AL and data visualization to create interactive AL algorithms involving data exploration and visualization, in a similar manner as in the annotation platform of \cite{kth_paper}.

\section{Acknowledgements}

This research was funded by Academy of Finland grants no. 314602, 335872, and 343498.

\bibliographystyle{IEEEtran}

\bibliography{mybib}

\end{document}